\documentclass{article}

\usepackage{arxiv}
\usepackage[utf8]{inputenc} 
\usepackage[T1]{fontenc}    
\usepackage{hyperref}       
\usepackage{url}            
\usepackage{booktabs}       
\usepackage{amsfonts}       
\usepackage{nicefrac}       
\usepackage{microtype}      
\usepackage{lipsum}
\usepackage{graphicx}

\title{Symmetrical Gaussian Error Linear Units (SGELUs) }

\author{
  Chao ~Yu \\
  State Key Laboratory of Millimeter Waves\\
  Southeast University\\
  Nanjing, 210096 \\
  \texttt{chao.yu@seu.edu.cn} \\
   \And
 Zhiguo~Su \\
  Jiangsu Smartwin Electronics Technology Co., Ltd.\\
  Jurong, Jiangsu Province, 212400\\
  \\
  \texttt{scott\textunderscore su\textunderscore phd@163.com} \\
}

\begin{document}
\maketitle

\begin{abstract}
In this paper, a novel neural network activation function, called Symmetrical Gaussian Error Linear Unit (SGELU), is proposed to obtain high performance. It is achieved by effectively integrating the property of the stochastic regularizer in the Gaussian Error Linear Unit (GELU) with the symmetrical characteristics. Combining with these two merits, the proposed unit introduces the capability of the bi-direction convergence to successfully optimize the network without the gradient diminishing problem. The evaluations of SGELU against GELU and Linearly Scaled Hyperbolic Tangent (LiSHT) have been carried out on MNIST classification and MNIST auto-encoder, which provide great validations in terms of the performance, the convergence rate among these applications.
\end{abstract}


\section{Introduction}

Artificial intelligent (AI) has shown its power in changing the world in a smart way, such as the well-known AlphaZero \cite{Silver1140}. To accomplish complicated tasks, deep neural network in which the activation functions play an important role in achieving the goal has been employed. In 2010, Rectified Linear Units (ReLUs) were proposed to realize better object recognition and face verification  \cite{nair2010rectified}, which has been extensively applied in daily life. To further enhance the capability of ReLUs, Exponential Linear Units (ELUs) were developed in 2015 \cite{clevert2015fast}. After that, a sigmoid-weighted linear unit called SWISH was found to realize a smooth, non-monotonic function \cite{ramachandran2017swish}. To produce better performance, Gaussian error linear units (GELUs) were proposed to include the stochastic distribution depending on the input\cite{hendrycks2016gaussian}.However, most of the activation functions suffer from the gradient diminishing problem. To resolve this issue, a non-parametric Linearily Scaled Hyperbolic Tangent (LiSHT) activation function was proposed by introducing the symmetrical structure with good performance \cite{swalpa2019lisht}.

To further improve the task performance, we propose a novel activation function in this work, named Symmetrical Gaussian Error Linear Units (SGELUs), which provides extraordinary merits by introducing the symmetrical characteristics into GELUs. Firstly, it can successfully employ the negative value to resolve the gradient diminishing problem. Secondly, it can effectively utilize the stochastic characteristics originated from GELUs to weigh the input. By these operations, the convergence and accuracy can be further improved against GELUs and LiSHT, which has been validated in the tasks, such as the MNIST classification and the MNIST auto-encoder.

\section{SGELU Formulation}

It is widely recognized that the activation function of GELU can be represented by

\begin{equation}
GELU(x)=xP(X\leq x)= \frac{1}{2}x(1+erf(\frac{x}{\sqrt2}) )
  \label{equ:equ1}
\end{equation}

\noindent where $erf(\cdot)$ represents the Gaussion error function, that is,

\begin{equation}
erf(x)=\frac{2}{\sqrt{\pi}}\int_{0}^{x}e^{-t^{2}}\, dt
\label{equ:equ2}
\end{equation}

Since the GELU function represents the nonlinearity using the stochastic regularizer on an input, which is the cumulative distribution function derived from the Gaussian error function, it has shown the advantage over other functions, e.g., ReLU, ELU. However, most activation functions do not fully exploit the negative value. Taking this into account, we take the advantage of stochastic regularizer on the input and exploit the negative value, and propose a novel Symmetrical Gaussian Error Linear Unit (SGELU), which can be represented by

\begin{equation}
SGELU(x)=\alpha x*erf(\frac{x}{\sqrt2})
\label{equ:equ3}
\end{equation}

\noindent in which $\alpha$ represents the hyper-parameter that can be tuned in the computation to obtain the optimum solution.

\begin{figure}
  \centering
  \includegraphics[width=0.8\columnwidth]{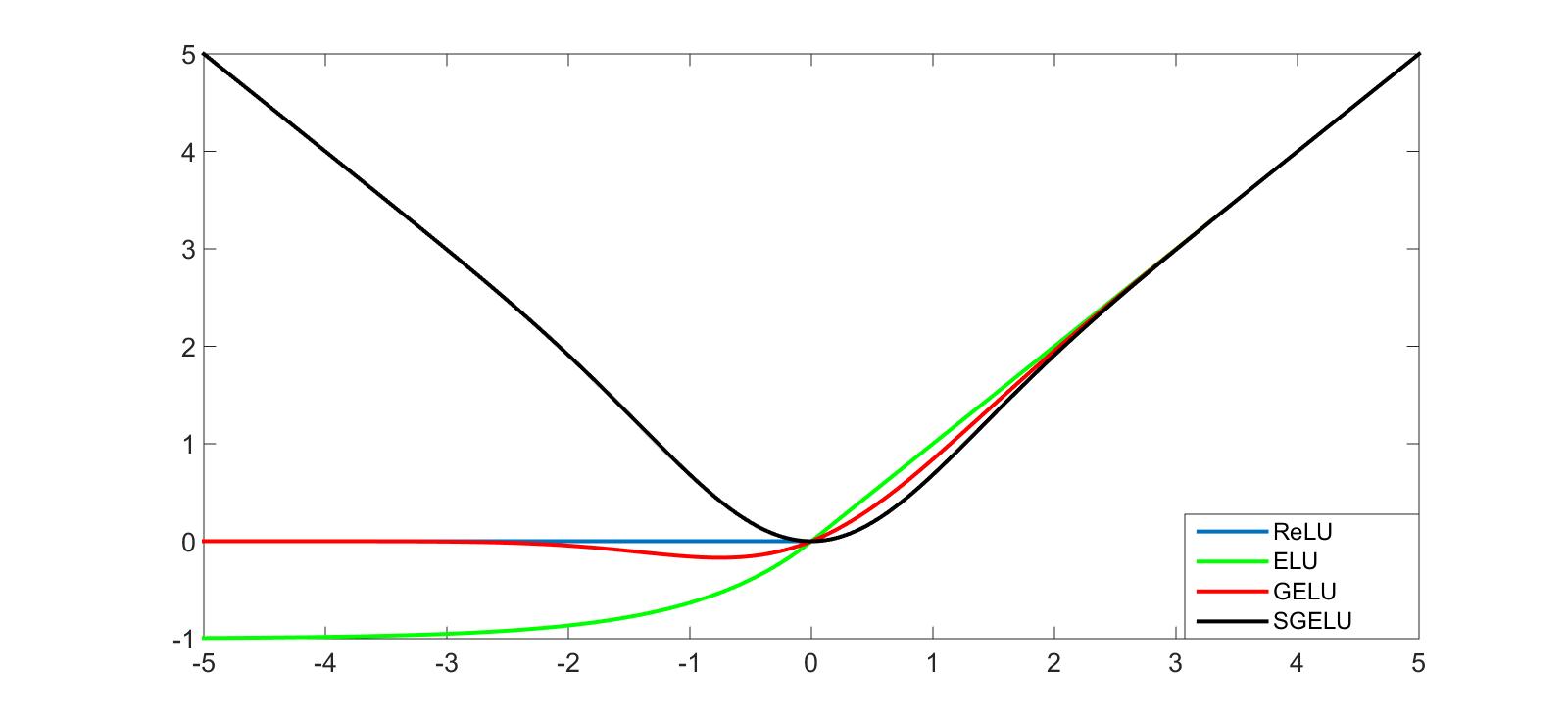}
  \caption{The SGELU, GELU, ReLU and ELU.}
  \label{fig:Figure1}
\end{figure}

The properties of different activation functions have been shown in Figure \ref{fig:Figure1}. In Figure \ref{fig:Figure1}, the proposed SGELU activation function is different from GELU, ReLU and ELU, which shows the symmetrical characteristics together with Gaussian regularizer. To further understand the symmetrical property, the gradient function and weights updating process of SGELU are compared with those of ReLU, ELU and GELU. Since the gradient function of SWISH \cite{ramachandran2017swish} is similar to that of GELU, it is not included here.

For simplification, the loss function for the single layer in deep neural network is defined as follows,

\begin{equation}
E=\frac{1}{2}(\hat{y}-y)^2
\label{equ:equ4}
\end{equation}

where $\hat{y}$ and $y$ are the true value and the predicted value, respectively.
\begin{equation}
y=f(z)
\label{equ:equ5}
\end{equation}

where $f(\cdot)$ is the activation function, such as ReLU, ELU, GELU or SGELU, and

\begin{equation}
z = wx+b
\label{equ:equ6}
\end{equation}

\noindent where $w$ is the weight and $b$ is the bias. The input $x$ and the output $y$ are all normalized to the interval of [0, 1]. Then  the gradient function can be easily derived according to the chain rule, that is,

\begin{equation}
\frac{\partial E}{\partial w}=-(\hat{y}-y)\frac{\partial f(z)}{\partial z} x
\label{equ:equ7}
\end{equation}

Finally, the value of the weight can be updated by the following equation:

\begin{equation}
w_{n+1}=w_n-\frac{\partial E}{\partial w}
\label{equ:equ8}
\end{equation}

\begin{figure}
  \centering
  \includegraphics[width=0.8\columnwidth]{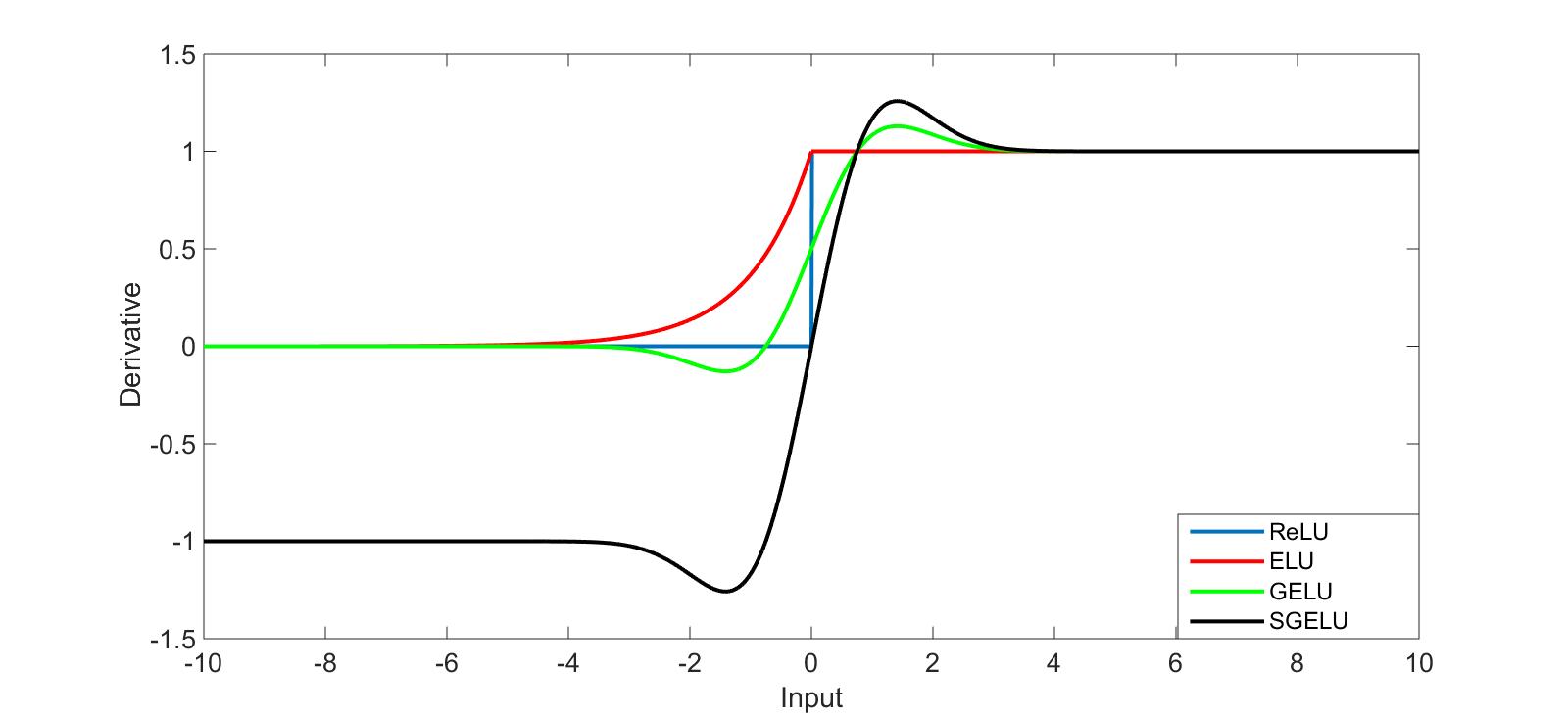}
  \caption{Derivatives of SGELU, GELU, ReLU and ELU.}
  \label{fig:Figure2}
\end{figure}

For better illustration, the derivatives of ReLU, ELU, GELU and SGELU are plotted as shown in Figure \ref{fig:Figure2}, corresponding to those activation functions in Figure \ref{fig:Figure1}. To analyze the weight updating process in terms of different activation functions, we assume that the current predicted output $y$ of the network using different activation functions is smaller than the true value of $\hat{y}$.

(a)	For ReLU \\
If the current input $z$ is positive, $\frac{\partial f(z)}{\partial z} $ is positive. Then, the value in gradient function (\ref{equ:equ7}) is negative, resulting that the weight increases according to (\ref{equ:equ8}), which makes the output increase and approach the real value. But if $z$ is negative, the gradient is zero and thus the weight stops updating.

(b)	For ELU \\
The weight updating of ELU is similar to the one of ReLU, when $z$ is positive. But if  $z$ is negative, the $\frac{\partial f(z)}{\partial z} $ is positive but with small values. The gradient in (\ref{equ:equ7}) is negative, so that the weight updates up to a bigger value and moves towards to the positive direction with a relatively slow learning rate.

(c)	For GELU \\
If $z$ is positive, the situation of GELU is similar to the one of ReLU or ELU. However, if  $z$ is negative, the most of $\frac{\partial f(z)}{\partial z} $  is negative. And the gradient value in (\ref{equ:equ7}) is then positive or very close to zero if $z$ is “very” negative for most cases, which pushes weight to a smaller value. Finally, the weight stops updating. The resulting output is deviating from the true value until the weight stops updating.

(d)	For SGELU \\
The similar procedure happens for SGELU when $z$ is positive, compared with GELU. However, it is different when $z$ is negative, that is, the gradient value in (\ref{equ:equ7}) is positive. The weight becomes smaller according to (\ref{equ:equ8}), but the output increases, due to the symmetrical characteristics about y-axis of the SGELU function.

From the discussion above, the biggest difference of SGELU from other activation functions is what happens in the negative half axis. Instead of forcing the output to be zero like ReLU, or deviating the output from true value like GELU until it stops converging, or dragging the output from negative to positive like ELU, SGELU can update its weight symmetrically towards to two directions in both positive and negative half axis. In other words, the function of SGELU is a two-to-one mapping between the input and the output, while the others are a one-to-one mapping. Meanwhile, different from LiSHT, the Gaussian stochastic characteristics is employed to weight the input, resulting further enhancement of the model capability. Therefore, these features of SGELU makes it more efficient and robust.

\section{SGELU Experiments}

To validate the proposed idea, we evaluate SGELU, GELU and LiSHT on MNIST classification, MNIST auto-encoding. The performance for classical activation functions such as ReLU, and ELU is not included here, since GELU outperforms them which has been validated in \cite{hendrycks2016gaussian}.

\subsection{MNIST Classification}
\begin{figure}
  \centering
  \includegraphics[width=0.8\columnwidth]{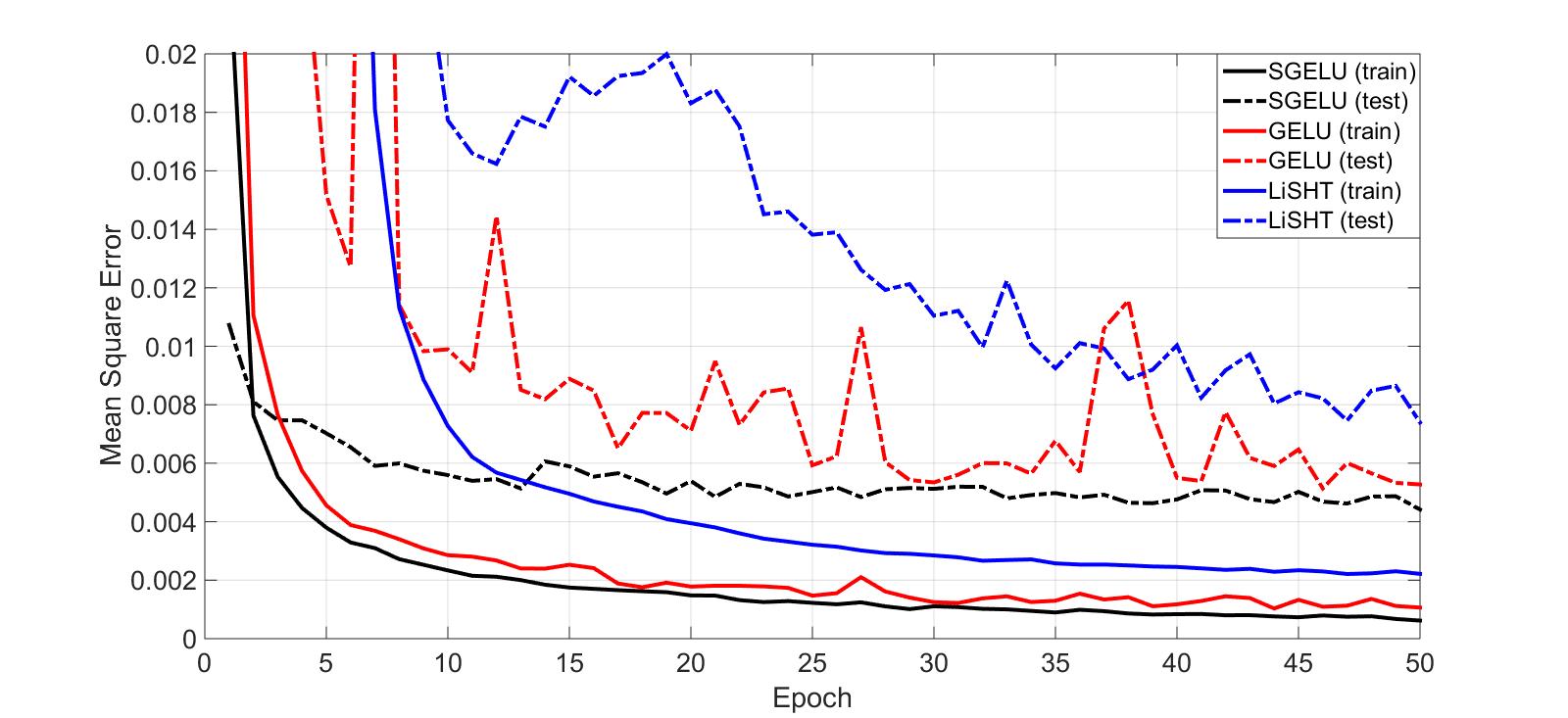}
  \caption{MNIST Classification Results. Train results are represented by solid lines, and test results are represented by broken lines.}
  \label{fig:Figure3}
\end{figure}

In this validation, a fully connected SGELU neural network with $\alpha$=0.1 is trained to compare with a similar network using GELU and LiSHT. Similar to \cite{hendrycks2016gaussian}, each 8-layer, 128-neuron wide neural network is trained for 50 epochs with a batch size of 128, in which the Adam optimizer is employed \cite{kingma2014adam}. Different from \cite{hendrycks2016gaussian}, the loss function defined here is mean square error (MSE). Training data, 51,200 samples (400 batches), are extracted from MNIST training set, and 6,400 samples (50 batches) from MNIST test set are used to test trained networks. All three networks are trained and tested five times, and the comparison of the median results for five runs are presented in Figure \ref{fig:Figure3}, which demonstrates SGELU is more accurate than GELU and LiSHT. As shown in Figure \ref{fig:Figure3}, the out-of-sample error (test) of the network using SGELU (black and black broken lines), comparing with the network using GELU (red and red broken line) and LiSHT (blue and blue broken line), is less and decreases more smoothly, which means the generalization of the network using SGELU is better.

It is worth mentioning that the batch normalization (BN) \cite{ioffe2015batch} technique is deployed for GELU network before the GELU activation function in every neuron. However, in the proposed the SGELU network, instead of employing BN, a simple min-max normalization technique is performed after every output of the SGELU activation function for each batch individually, except the last layer (the output layer). This modification can reduce the training time significantly, because the computational complexity of min-max normalization is much less than that of BN.

\subsection{ MNIST Auto-encoder}
\begin{figure}
  \centering
  \includegraphics[width=0.8\columnwidth]{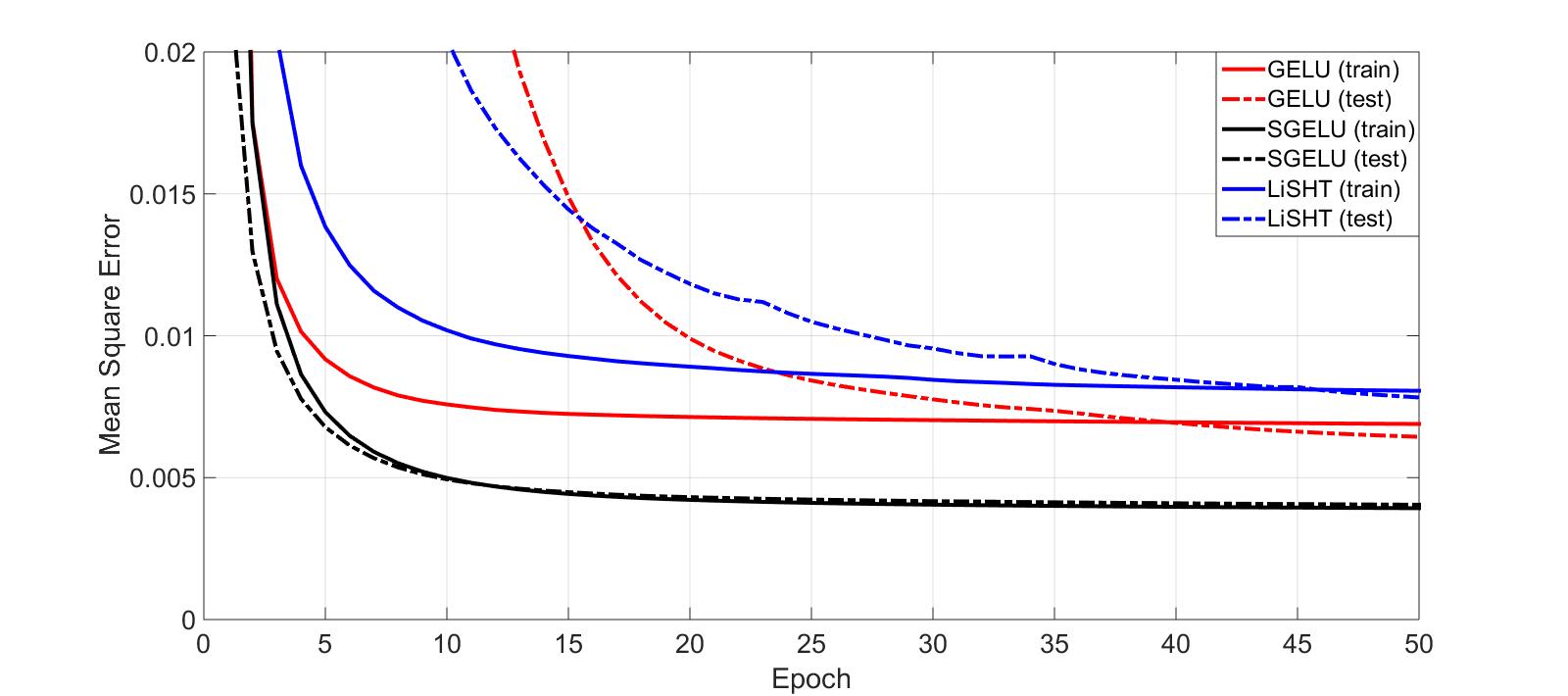}
  \caption{MNIST AutoEncoder Results. Train results are represented by solid lines and test results are represented by broken lines.}
  \label{fig:Figure4}
\end{figure}

To compare the performance in the task of auto-encoder, a simply network with only one encoder layer and one decoder layer is constructed and trained, in which 128 neurons is chosen for each layer. The reason for employing the swallow network is that 1)it is less time-consuming to train small newtork, and 2) auto-encoder network is usually trained layer by layer individually, e.g., stacked auto-encoders. The batch size of 128 and the Adam optimizer are employed here again. The hyper-parameter $\alpha$ is set as 0.1 for SGELUs. The training data of 51,200 samples (400 batches) from MNIST train set and the test data of 6,400 samples (50 batches) from MNIST test set are employed in this demonstration. BN and min-max normalization are respectively equipped with GELU/LiSHT and SGELU auto-encoder networks. The training and the testing median results for five runs are plotted in Figure \ref{fig:Figure4}, which shows that SGELU significantly outperforms GELU and LiSHT.

\section{Discussion}

\subsection{Weigh Distribution}

\begin{figure}
  \centering
  \includegraphics[width=0.8\columnwidth]{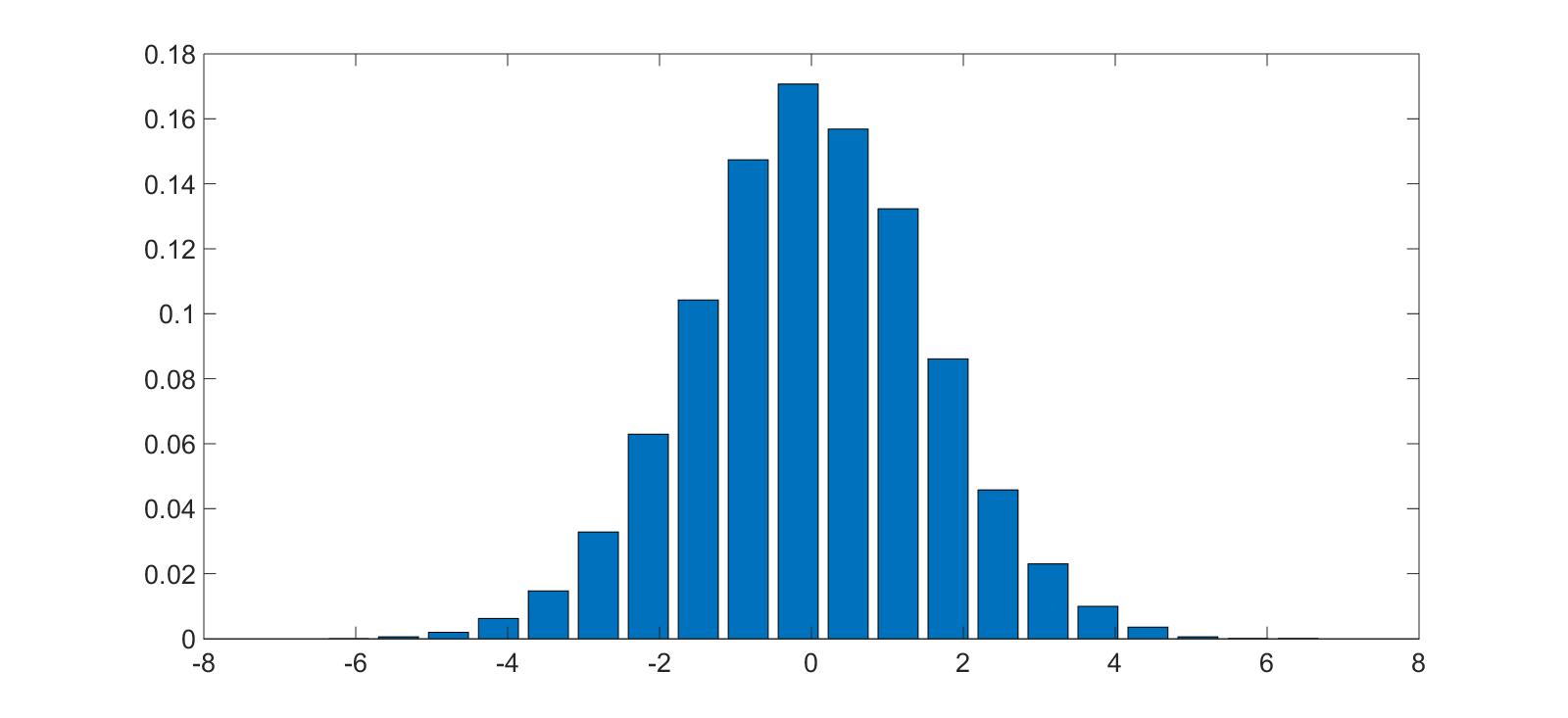}
  \caption{Histogram of the SGELU weights.}
  \label{fig:Figure5}
\end{figure}

As discussed above, the weights of SGELU are more likely to “spread” symmetrically as shown in Figure \ref{fig:Figure5}.
To validate the pattern, the weights of SGELU and GELU in MNIST classification experiment are inspected using Kolmogorov-Smirnov normal distribution test \cite{brillinger1967handbook}, and the weights from hidden layer 1 to 8 (except the input and output layers) for both networks are tested. Through the tests, the weights in seven layers of the SGELU network passed the test as the normal distribution, but only the weights in two layers of the GELU network passed the test. Generally, normal distribution of weights can make network more rational, accurate, and robust. More precisely, in order to represent the relationship between input and output, big values of the weights are likely to realize the coarse and linear mapping, and small values of the weights are responsible for fine and nonlinear mapping. In these operations, the normal distribution of SGELU is crucial for achieving great performance.

\subsection{Dropout}

The dropout operation is also studied in the SGELU network. However, the SGELU does not comport well with this operation. The possible reason is that the ReLU and GELU force the function at negative x-axis to output zeros or small values, which is sparse and consequently comports well with dropout. Different from ReLU, GELU and ELU, SGELU utilizes the outputs in the negative x-axis, which avoids neurons to be dead. Therefore, it is not necessary to employ dropout in the SGELU network.

\subsection{Batch Normalization }
In the SGELU network, if the BN is employed before the activation functions, the performance is not as good as the one with a min-max normalization utilized after the activation functions, but it can still achieve similar accuracy compared to the GELU network. The reason is that the SGELU function does not need the input to become the normal distribution which could limit SGELU to only employ its fine and nonlinear mapping part to fit output. However, it is still required to employ SGELU’s coarse and linear mapping part. Besides, the min-max normalization used in SEGLU network is much less computational complexity comparing with BN, because there are no any mean, variance, and moving average computation involved.

\section{Conclusion}
In this paper, a novel activation function SGELU is proposed to effectively enhance the model accuracy. Compared to GELU, SGELU inherits its stochastic regularizing feature and introduces the new symmetrical feature, leading to a two-to-one mapping and bi-directional optimization. Therefore, the proposed SGELU can be an excellent alternative to other activation functions and provide a new guideline for searching novel activation functions.



\end{document}